\documentclass[manuscript,screen]{acmart}
\AtBeginDocument{%
  \providecommand\BibTeX{{%
    \normalfont B\kern-0.5em{\scshape i\kern-0.25em b}\kern-0.8em\TeX}}}

\setcopyright{rightsretained}
\copyrightyear{2023}
\acmYear{2023}
\acmDOI{}

\acmConference[CHI '23]{Generative AI and HCI}{April 23--28,
  2023}{Hamburg, Germany}
\acmBooktitle{CHI '23: ACM CHI Conference on Human Factors in Computing Systems,
 April 23--28, 2023, Hamburg, Germany} 
\acmPrice{15.00}
\acmISBN{978-1-4503-XXXX-X/18/06}

\usepackage{caption}
\usepackage{subcaption}
\usepackage{booktabs}
\usepackage{graphicx}
\usepackage{framed}

\newcommand{\nb}[3]{
  \fcolorbox{black}{#2}{\bfseries\sffamily\scriptsize#1}
    {\sf\small$\blacktriangleright$\textit{#3}$\blacktriangleleft$}
}

\newcommand\lore[1]{\nb{Lorenzo}{orange}{#1}}

\begin{document}

\title{ARTIST: ARTificial Intelligence for Simplified Text}

\author{Lorenzo Corti}
\orcid{0000-0002-4200-1664}
\email{l.corti@tudelft.nl}
\author{Jie Yang}
\orcid{0000-0002-0350-0313}
\email{j.yang-3@tudelft.nl}
\affiliation{%
  \institution{Delft University of Technology}
  \country{Netherlands}
}


\begin{abstract}
    Complex text is a major barrier for many citizens when accessing public information and knowledge.
    While often done manually, Text Simplification is a key Natural Language Processing task that aims for reducing the linguistic complexity of a text while preserving the original meaning.
    Recent advances in Generative Artificial Intelligence (AI) have enabled automatic text simplification both on the lexical and syntactical levels.
    However, as applications often focus on English, little is understood about the effectiveness of Generative AI techniques on low-resource languages such as Dutch.
    For this reason, we carry out empirical studies to understand the benefits and limitations of applying generative technologies for text simplification and provide the following outcomes: 
    \emph{1)} the design and implementation for a configurable text simplification pipeline that orchestrates state-of-the-art generative text simplification models, domain and reader adaptation, and visualisation modules; 
    \emph{2)} insights and lessons learned, showing the strengths of automatic text simplification while exposing the challenges in handling cultural and commonsense knowledge.
    These outcomes represent a first step in the exploration of Dutch text simplification and shed light on future endeavours both for research and practice.
\end{abstract}





\maketitle

\section{Introduction}
\label{sec:intro}

Literacy is defined as ``the ability to understand and respond appropriately to written texts'', and forms a key information processing competency \citep{oecd2019survey}.
Embedded clauses, passive voice, non-canonical word order, and low-frequency words can disrupt the sequential flow of sentences and arguments \citep{gasperin2009natural}.
In the Netherlands, about 2.5 million citizens between 16 and 65 years old are affected by low-literacy levels and find it challenging to participate in everyday society, e.g., when reading newspapers. \footnote{\url{https://www.lezenenschrijven.nl/informatie-over-laaggeletterdheid-nederland}}
As a recent example of such consequences, some citizens experienced difficulties in understanding official communications during the COVID-19 pandemic. \footnote{
\url{ https://www.rtlnieuws.nl/editienl/artikel/5100711/persconferentie-rutte-moeilijk-laaggeletterden-stichting-lezen-schrijven}
}
%
%
%
%
In an effort to address literacy gaps, a number of resources such as corpus-specific frequency lists \citep{bauman1995general}, controlled languages such as ASD-STE100\footnote{ASD-STE100: \url{https://asd-ste100.org/about.html}}, and guidelines for accessible communication \citep{ukgov2022accessible} have been developed to aid manual simplification.
%
Specific to the Netherlands, several initiatives have been undertaken to manually make books or news articles accessible \footnote{See
\url{https://www.nieuwslezer.bibliotheek.nl/}}.
However, this is not trivially achieved and is labour-intensive --- irrespective of the language ---, making a small selection of publications accessible.

Text Simplification (TS) has been a long-standing task within Natural Language Processing (NLP) and has been attracting continuous interest.
In the last few decades, methods for both lexical and syntactic simplification have been developed; from rule-based to data-driven approaches \citep{glavavs2015simplifying, horn2014learning, specia2010translating, wang2016experimental, zhang2017sentence}.
Particularly, generative (large) language models have progressed considerably and can now produce human-like simplifications that avoid long, complex, and conjoined sentences.
These models can capture the distributional semantics of words \citep{mikolov2013distributed, devlin2018bert}, a key requirement for meaning preservation and text coherence.
However, despite their success, these models are still at an early stage and are unreliable often capturing spurious or biased correlations. Such a problem is especially pronounced in low-resource languages, such as Dutch.
Indeed, limited research and resources for Dutch text simplification are available \citep{bulte2018automating, vandeghinste2019wablieft, pander2014t, maks2005eferentie, rbn2014referentie}.
%

In this paper, we explore the application of generative AI for automatic text simplification.
Particularly, we focus on Dutch and assess \emph{candidate} text simplifications --- which can be later adapted by human experts --- both from quantitative and qualitative perspectives.
We showcase ARTIST -- ARTificial Intelligence for Simplified Text, a generative text simplification pipeline integrating readability assessments, of which we release the implementation. \footnote{Code repository for ARTIST: \url{https://github.com/delftcrowd/ARTIST}}
We conclude by summarising the lessons learned and by providing an outlook for future research on generative text simplification.
%
\section{ARTIST} \label{sec:artist}

\begin{figure}[t]
     \centering
     \begin{subfigure}[b]{0.475\textwidth}
         \centering
         \includegraphics[width=\textwidth]{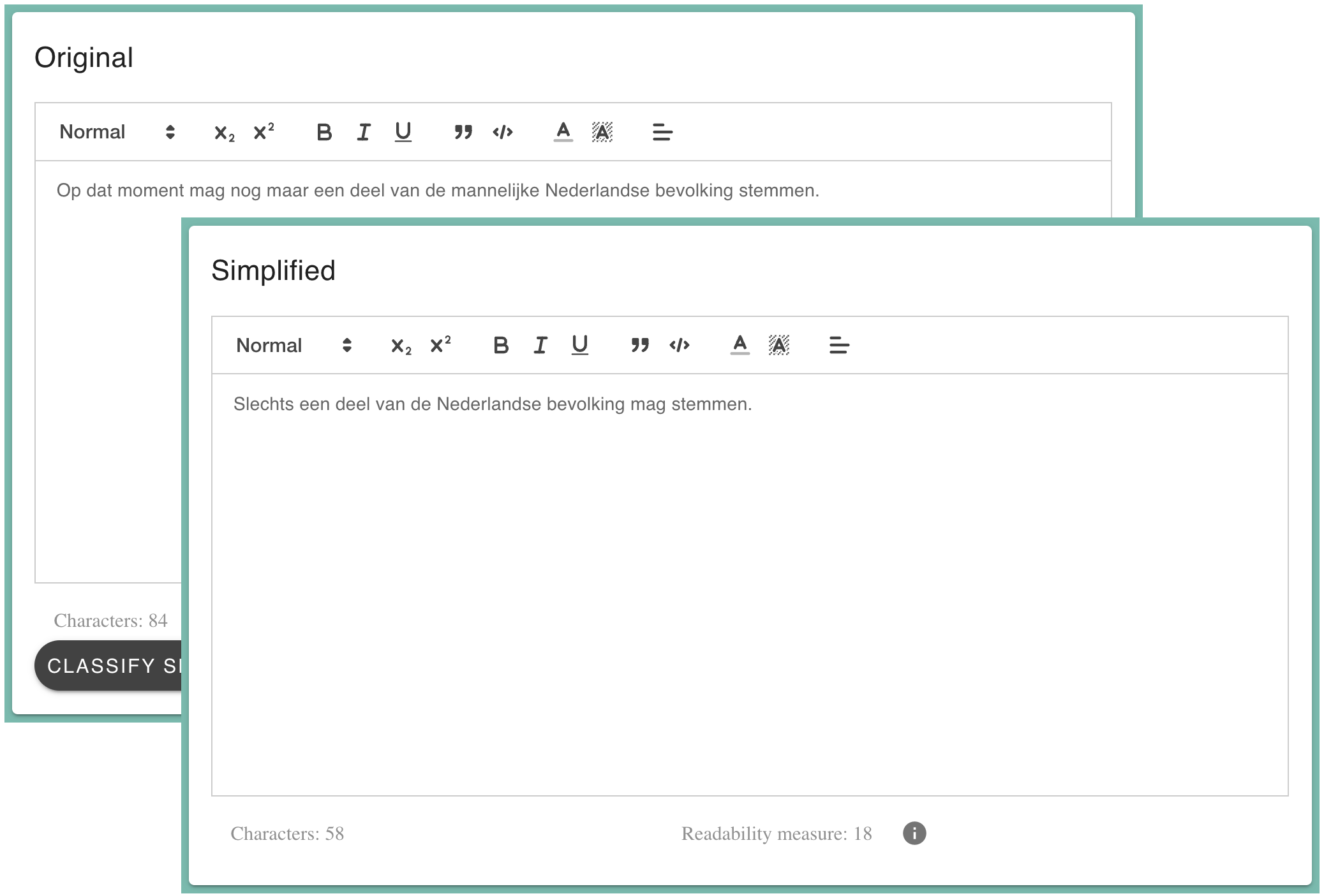}
         \caption{Simplification User Interface.}
         \label{fig:UI1}
     \end{subfigure}
     \hfill
     \begin{subfigure}[b]{0.475\textwidth}
         \centering
         \includegraphics[width=\textwidth]{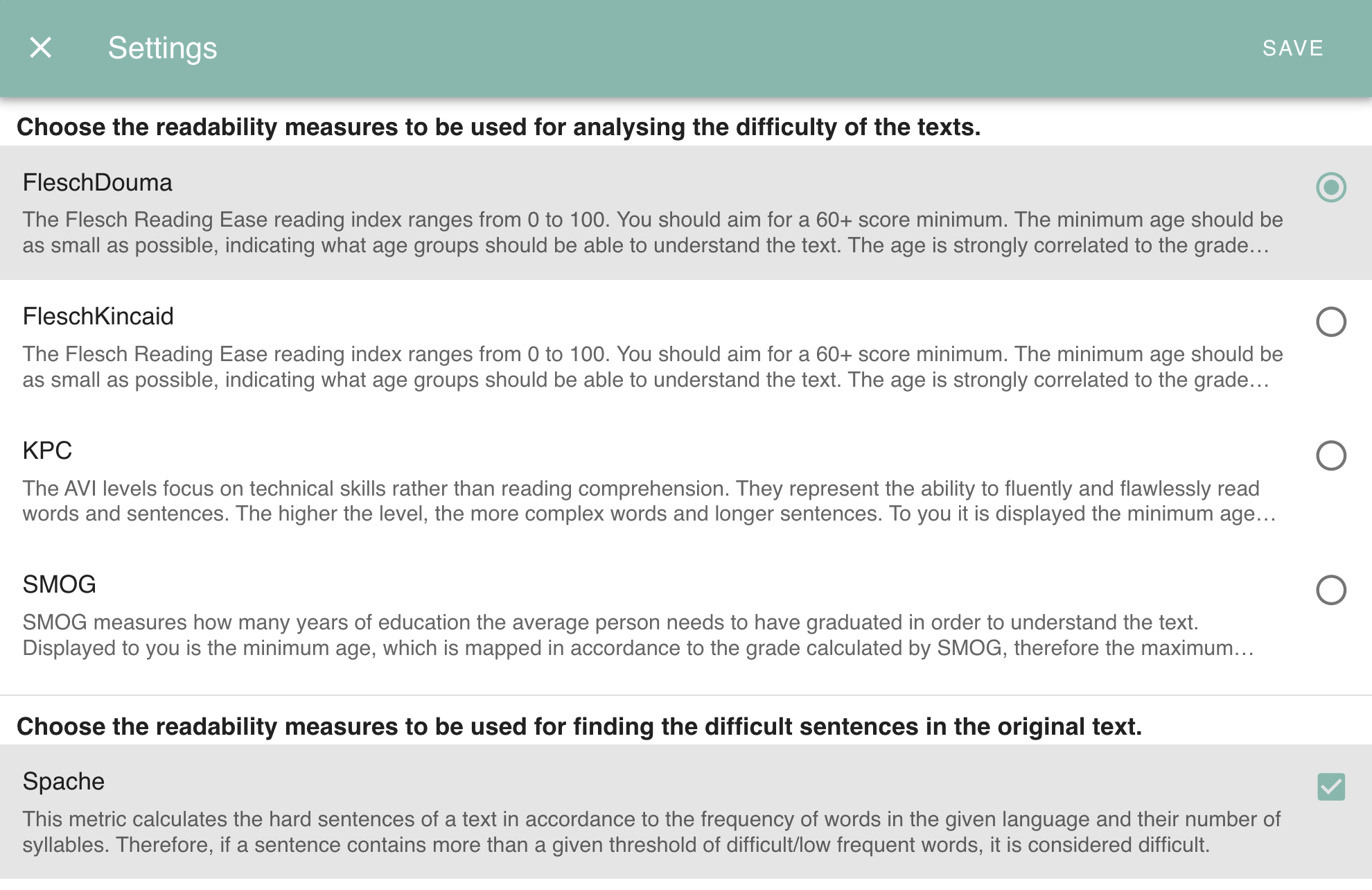}
         \caption{User-configurable settings.}
         \label{fig:UI2}
     \end{subfigure}
     \caption{ARTIST's ext simplification user interface (a) and settings (b).}
     \label{fig:artist_ui}
\end{figure}

In this section, we present the implementation for ARTIST, a generative text simplification pipeline.
ARTIST takes the shape of a web application (Figure \ref{fig:artist_ui}) which combines state-of-the-art text simplification models and readability assessments while allowing for user configurability.
Model-wise, we experiment with two configurations, both of which leverage the Text-to-Text Transfer Transformer (T5) model \citep{raffel2020exploring}. For the former, we use T5 fine-tuned on Dutch CNN news. \footnote{See \url{https://huggingface.co/yhavinga/t5-v1.1-base-dutch-cnn-test}}
Instead, for the latter, we use a combination of translation and summarisation: we first translate \emph{complex} Dutch text to English \footnote{See \url{https://pypi.org/project/googletrans/}}, summarise it using the TS\_T5 model \cite{sheang2021controllable}, and then we translate it back to Dutch.
%
In the remainder of the paper, we will refer to these two configurations as \texttt{Dutch\_T5} and \texttt{GoogleTransl} respectively.
Regarding readability assessments, there are a number of metrics which are --- probably unsurprisingly --- specific to English and should be taken with caution.
Within ARTIST, we offer users the choice between Flesch Kincaid \citep{kincaid1975derivation}, Flesch Duoma \citep{douma1960leesbaarheid}, SARI \citep{xu2016optimizing}, SMOG \citep{mc1969smog}, and KPC \footnote{KPC method for computing AVI levels: \url{https://nl.wikipedia.org/wiki/AVI_\%28onderwijs\%29}} for text-level readability and Spache \citep{spache1974spache} for sentence-level readability.
%
%

\section{Results and Evaluation}
\label{sec:results}
We provide a brief overview of ARTIST both from a quantitative and qualitative standpoint.
To do so, we use the dataset from \emph{Canon van Nederland} (CvN) \footnote{Canon van Nederland: \url{https://www.canonvannederland.nl/}. Data was used with permission.}.
The CvN dataset is a collection of 50 topics that provide a chronological summary of Dutch history, meant to be taught in primary and secondary schools in the Netherlands.
For our experiments, we considered texts written for the upper level of secondary education and manually aligned $\langle$ simple; complex $\rangle$ text pairs.


\subsection{Quantitative evaluation}
We measure the performance of the two simplification strategies (\texttt{Dutch\_T5} and \texttt{GoogleTransl}) in terms of BLEU \citep{papineni2002bleu} --- a metric borrowed from machine translation --- by comparing the automatically simplified samples with their corresponding manually simplified variant.
Specifically, such reference samples from Canon van Nederland were manually simplified to be understandable by readers with lower-level secondary education (corresponding to 5th to 8th grade).
Table \ref{tab:bleu_scores} lists the 5 articles for which we obtain the highest BLEU scores for both models.
Overall, we obtain relatively low scores (BLEU scores range between 0 and 1) which may indicate lacklustre text simplifications.
However, instead of applying targeted edits, generative AI models re-create text which can negatively impact BLEU measurements.
Despite its shortcomings \citep{xu2016optimizing}, BLEU was found to correlate with human judgement \citep{stajner-etal-2014-one}.
\begin{table}[t]
    \caption{Quantitative evaluation for the analysed text simplification strategies.}
    \begin{subtable}[h]{0.35\textwidth}
        \centering
        \caption{Top-5 scoring articles from CvN dataset.}
        \begin{tabular}{@{}ll|ll@{}}
        \toprule
        \multicolumn{2}{c|}{\texttt{Dutch\_T5}} & \multicolumn{2}{c}{\texttt{GoogleTransl}} \\ \midrule
        Anton de Kom         & 0.138   & Anton de Kom              & 0.080  \\
        Jeroen Bosch         & 0.117   & Annie M.G. Schmidt        & 0.072  \\
        Trijntje             & 0.099   & De Atlas Maior van Blaeu  & 0.064  \\
        Hebban olle vogala   & 0.086   & Jeroen Bosch              & 0.063  \\
        Napoleon Bonaparte   & 0.086   & Aletta Jacobs             & 0.057  \\ \bottomrule
        \end{tabular}%
        \label{tab:bleu_scores}
    \end{subtable}
    \hfill
    \begin{subtable}[h]{0.4\textwidth}
        \caption{Manual evaluation of the automatically simplified text.}
        \scalebox{.9}{
            \begin{tabular}{@{}lrrr@{}}
            \toprule
            Model          & Simplicity & Fluency & Adequacy \\ \midrule
            \texttt{Dutch\_T5}      & 1.9        & 2.0     & 1.9      \\
            \texttt{GoogleTransl} & 1.5        & 1.3     & 1.8      \\ \bottomrule
            \end{tabular}%
        }
        \label{tab:human_eval}
     \end{subtable}
     \label{tab:temps}
\end{table}
%
%
%
Motivated by this, we complement the automatic evaluation with a manual analysis, of selected samples, in terms of lexical \textbf{simplicity}, sentence \textbf{fluency}, and \textbf{adequacy} with respect to the original text.
We chose four topics for manual evaluation, for each model the two topics with the highest BLEU score and two with the lowest BLEU score.
%
%
%
Each topic was judged by two human raters on a 5-point scale.
We observe that, although \texttt{Dutch\_T5} scores slightly higher than \texttt{GoogleTransl}, both received poor ratings, signalling room for improvement with respect to simplicity, fluency, and adequacy (see Table \ref{tab:human_eval}).
%
%

\subsection{Qualitative evaluation}
Given the poor results on automatic metrics, we qualitatively assess specific examples to illustrate where the models perform well and where they do not. 
We provide both the Dutch text and English translation for clarity.
\par\textbf{Simplifications can be aggressive} and important details might be obfuscated. Following is an example:
\begin{quote}
    \emph{Op dat moment mag nog maar een deel van de mannelijke Nederlandse bevolking stemmen.}
    
    \noindent\textbf{EN}: At that time, only part of the male population is allowed to vote. 
\end{quote}
While the original text emphasises that only men are allowed to vote, in the simplified text such detail is obfuscated and generalised:
\begin{quote}
    \emph{Slechts een deel van de Nederlandse bevolking mag stemmen.}
    
    \noindent\textbf{EN}: Only part of the population is allowed to vote.
\end{quote}
%
\par\textbf{Temporal consistency} is not guaranteed and dates might be changed.
For example:
\begin{quote}
    \emph{... van het 18e Internationale Vrouwencongres in 1915.}
    
    \noindent\textbf{EN}: ... of the 18th International Women's Congress in 1915.
\end{quote}
%
is incorrectly simplified to:
\begin{quote}
    \emph{... van het 18e Internationale Vrouwencongres in 2015.}
    
    \noindent\textbf{EN}: ... of the 18th International Women's Congress in 2015.
\end{quote}
%

\par\textbf{Factual correctness} is not guaranteed and facts might be merged into false ones. For example:
%
\begin{quote}
    \emph{Christiaan Huygens wordt in 1629 geboren als tweede zoon van Suzanna van Baerle en Constantijn Huygens, dichter en secretaris van twee prinsen van Oranje.}
    
    \noindent\textbf{EN}: Christiaan Huygens was born in 1629 as the second son of Suzanna van Baerle and Constantijn Huygens, a poet and secretary of the Princes of Orange.
\end{quote}
%

Besides merging facts, this simplification refers to Christiaan Huygens as `he' right from the beginning:

\begin{quote}
    \emph{Hij wordt geboren als tweede zoon van Suzanna van Baerle. Hij was secretaris van twee prinsen van Oranje.} 
    
    \noindent\textbf{EN}: He was born as the second son of Suzanna van Baerle. He was secretary of two Princes of Orange.
\end{quote}

\section{Discussion} \label{sec:discussion}

Our exploration of generative text simplification reveals its potential for low-resource languages like Dutch. However, several challenges still stand and can only be tackled with joint research efforts from related disciplines. 
In particular, we notice that the lack of world knowledge is an important issue of machine-learned TS models.
Those models make incorrect inferences about the relationships between entities
, leave out important details
, and create `alternative truths’ or falsehoods.
Furthermore, simplifications might be sensitive to domain and genre, thus making general-purpose simplifiers inadequate.
We acknowledge that, thus far, we have not included user studies in our experiments.
However, there are a number of considerations that can be made from the perspectives of end-users and language as different causes call for different TS solutions.
For example, reading levels (readability scores, age of acquisition, etc.) are largely ignored and are usually applied `as-is' from English.
For non-native speakers with a mastery of their mother tongue, reading is complicated by vocabulary, linguistic structure and discourse style \citep{candido2009supporting}.
On the other hand, for readers in a state of cognitive development, it is desirable to have a vocabulary which can be phonetically sounded and decoded to form sound-symbol correspondences.
Similarly, prepositions, words denoting sequences, and classifications (e.g., of animals) are essential and should be retained.
Furthermore, cognitive impairments (e.g., dyslexia) require alternative output layouts (typefaces, glossaries, pictures, etc.). 
%
%
To address these concerns, we foresee a number of research directions:
\begin{itemize}
    \item domain-specific adaptation of TS models
    \item neuro-symbolic view on integrating knowledge in TS models
    \item human-machine TS pipelines, where humans complement the weaknesses of automatic models
\end{itemize}
We conclude this paper by calling for collaboration across different disciplines (NLP, knowledge management, human computation, and human-machine interaction) to advance the current state of automatic text simplification solutions.


\begin{acks}
We would like to thank Cosming Anton, Pratham Johari, Marilotte Koning, Daniël Poolman, Kostas Stefanopoulos, and Beatrice Vizuroiu for helping us build the prototype for ARTIST.
This work was partially supported by the TU Delft Design@Scale AI Lab and done in collaboration with the Royal Library of the Netherlands.
In our project, we made use of the Dutch national e-infrastructure with the support of the SURF Cooperative using grant no. EINF-3132.
\end{acks}

\bibliographystyle{ACM-Reference-Format}
\bibliography{main}

\end{document}